# Optimal Path Planning for Aerial Load Transportation in Complex Environments using PSO-Improved Artificial Potential Fields

1st Ali Akbar Rezaei Lori
*Department of Mechanical and Material Engineering*
*University of Nebraska-Lincoln*
Lincoln, USA
arezaeilori2@huskers.unl.edu

*Abstract*— In this article, we investigate the optimal path planning for aerial load transportation in complex, dynamic, and static environments using Particle Swarm Optimization (PSO). A hierarchical optimal control system is designed for a quadrotor equipped with a cable-suspended payload, employing Euler-Lagrange equations of motion. To navigate through obstacles, an improved artificial potential field combined with the PSO algorithm is used to determine the shortest path for a virtual point, acting as a leader. This leader guides the system toward the target point while avoiding collisions with both fixed and moving obstacles. The gravitational and repulsion coefficient forces using various PSO methods are fine-tuned to achieve the best trajectory and minimize time duration. The identified point serves as the desired location for quadrotor position control, based on a sliding mode strategy. Finally, we present numerical results to demonstrate the successful transportation of the payload by the system.

## I. Introduction

In the past few years, there has been growing interest in the study of airborne payload transportation as a means to address issues like congested traffic and remote accessibility. Among the various types of aerial vehicles, quadrotors have garnered significant attention due to their broad capabilities. Their exceptional features, including easy vertical take-off and landing, mechanical simplicity, and outstanding maneuverability, have made them highly sought after.

At the early works, like [1], the motion equations of aerial system along with the suspended load in 2D coordinates where the sliding mode and feedback linearization control were used for transitional and rotational movements, respectively. Then, the aerial load transportation included a quadrotor and a slung load were considered as a coupled system in the most of researches, and different control methods were applied to control the system. For instance, motion equations of a quadrotor with a slung load were obtained by the Lagrange method in [2], and an anti-swing control algorithm was designed to reduce the load fluctuations. In [3], the position and attitude of a quadrotor, while transporting a mass point, were controlled, respectively, by a feedback linearization controller and a neural network controller. In another investigation [4], a method for lifting a suspended payload was introduced, based on the relationship between the weight of the load, distance from the quadrotor, and cable length. This approach divided the lifting procedure into three phases: preparation, pulling, and raising. When the distance between the load and the quadrotor was less than the cable length, the load's impact was disregarded. To account for the cable's elasticity in transporting a point mass, a parallel spring and damper system was employed [6].

In [7], researchers tackled the challenge of trajectory tracking for a quadrotor's linear dynamics model carrying a slung payload, using a mixed $H_2$ and $H_\infty$ control technique. In [8], an optimal-time trajectory control for transporting a point mass through aerial means was developed using the sequential quadratic programming (SQP) method. The investigation in [9] delved into swing-free trajectory tracking control for a quadrotor transporting a cable-suspended load, employing a dynamic programming algorithm for the control strategy. In [10], a novel approach combined the rotational and translational motion equations of a quadrotor carrying a suspended load in 2D coordinates through coordinate transformation. Subsequently, this new dynamic model was stabilized using an energy shaping control method. In [11], an optimal control approach was used for path planning to transport the maximum load mass with minimal swinging. Additionally, the optimal cable length was determined for carrying the maximum load mass. A robust control was applied in [12] to carry an unknown payload as well as manage the disturbances caused by wind forces and load swings.

Path planning for robotic systems in dense environments has recently become an important issue and some researches studied this task with different methods [13-15]. In applications of the UAVs, some methods has been used to observed the environment and avoid obstacles like the Conditional-Generative Adversarial Network in [16]. In the real world, payload transportation occurs in environments filled with obstacles. Some research has investigated obstacle avoidance in aerial load transportation, as seen in [17] and [18]. However, these studies primarily focused on obstacle avoidance assuming the presence of only static obstacles. In reality, operations take place in crowded environments with both static obstacles like buildings and dynamic obstacles like birds. Therefore, it is crucial to design a control system that can avoid collisions with both dynamic and static obstacles. Furthermore, factors such as energy consumption and operation duration in emergency cases are critical, as choosing a longer path instead of a shorter one can lead to problems.

The controller must address these issues by selecting the closest possible trajectory to reach the desired location while navigating through obstacles. Indeed, optimal path planning for aerial load transportation in dense and dynamic environments is of paramount importance.

In this article, we propose an optimal path planning approach for aerial load transportation in a dense and dynamic environment. This approach combines the improved artificial potential field (IAPF) method with classic and adaptive particle swarm optimization techniques (PSO). First, a mathematical model for a payload connected to a quadrotor is presented (Section 2). The optimal control is designed in three steps. First, the IAPF method is proposed, which takes into account the desired payload position, the current payload position, and the positions of obstacles. It generates artificial repulsive and attractive forces to assist in obstacle avoidance (Section 3.A). Additionally, we apply the PSO techniques to determine the optimal path to the desired position by adjusting the coefficients of the repulsive and attractive forces of the IAPF. Next, a sliding mode position control system is developed for the quadrotor to track the virtual point (Section 3.B). The attitude control is achieved using a PID controller (Section 3.C). Finally, we conduct numerical simulations to demonstrate and compare the performance of the entire load transportation control system (Section 4).

## II. QUADROTOR SUSPENDED-LOAD DYNAMICAL MODELING

The aerial load transportation system involves connecting a point mass load to a quadrotor with mass denoted as $m_q \in \mathbb{R}$ and moment of inertia $J_q \in \mathbb{R}^{3 \times 3}$ using a series of components, as illustrated in Figure 1. The cable is conceptualized as a sequence of rigid segments, each with its own mass, with the final mass being the payload itself. The inertial frame is established by $\{e_1, e_2, e_3\}$, where $e_3 = [0 \ 0 \ 1]^T$ indicates the direction of gravity, and the body frame $\{b_1, b_2, b_3\}$ is affixed to the center of mass of the quadrotor.

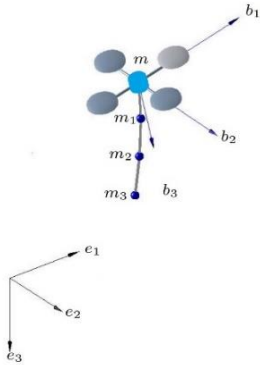

Figure 1: A quadrotor with a suspend load.

The spherical orthogonal group is define by $SO(3) = \{R \in \mathbb{R}^{3 \times 3} \mid R^T R = I, det[R] = 1\}$ to represent the orientation of the quadrotor, denoted as $R \in SO(3)$. The position of the quadrotor's center of mass and the length of the ith component are represented as $r_i \in \mathbb{R}^3$ and $l_i \in \mathbb{R}$, respectively. The direction of the *i*th component is denoted by $q_i = [q_{xi}, q_{yi}, q_{zi}] \in S^2$ such that $S^2 = \{q \in \mathbb{R}^3 \mid \|q\| = 1\}$.

Furthermore, the total thrust generated by the quadrotor is presented by $-fRe_3 \in \mathbb{R}^3$ in the inertial frame and the qyadrotor torque is describes by $\tau \in \mathbb{R}^3$ in its body frame, where, $-f \in \mathbb{R}$ is the magnitude of total thrust and $R \in \mathbb{R}^{3 \times 3}$ is the quadrotor rotational matrix. Additionally, the system is moving in the environment with unknown disturbances $F_{dis} \in \mathbb{R}^3$ affecting the quadrotor's translational movements the quadrotor translational movements and $\tau_{dis} \in \mathbb{R}^3$ t influencing its rotational motions.

The equations of motion are driven by [14] the Euler-Lagrange method as following:

$$M_T \ddot{r}_q - M_T g e_3 + \sum_{i=1}^{n} M_{qi} l_i \ddot{q}_i = -fRe_3 + F_{dis} \quad (1)$$

$$\sum_{j=1}^{n} M_{cij} l_i l_j \hat{q}_i^2 \ddot{q}_j + M_{qi} l_i \hat{q}_i^2 \ddot{r}_q + M_{qi} g l_j \hat{q}_i^2 e_3 = 0 \quad (2)$$

$$J_q \dot{\Omega}_q - \hat{\Omega}_q J_q \Omega_q = \tau + \tau_{dis}. \quad (3)$$

where $\omega_i \in \mathbb{R}^3$ and $\Omega_q \in \mathbb{R}^3$ are the angular velocity of the *i*th component and quadrotor angular velocity, respectively.

By using $\hat{q}_i^2 \ddot{q}_i = -\| \dot{q}_i \|^2 q_i - \ddot{q}_i$, and substitution in (1) to (3), we have

$$M\dot{X} = T - C \quad (4)$$

In here, $X = [\dot{r}_q \ \omega_1 \ \ldots \ \omega_n]^T \in \mathbb{R}^d$ is the state vector where the dimension $d = 3 + 3n$. Also, $M \in \mathbb{R}^{d \times d}$, $C \in \mathbb{R}^d$ and $T \in \mathbb{R}^d$ and corresponds to the mass matrix, Coriolis matrix and control input, respectively. The mass matrix is given as

$$M = \begin{bmatrix} M_T I_3 & M_{r_q \omega} \\ M_{\omega r_q} & M_\omega \end{bmatrix} \quad (5)$$

where each submatrix is defined by

$$M_{r_q \omega} = -[M_{q1} l_1 \hat{q}_1 \quad M_{q2} l_2 \hat{q}_2 \quad \ldots \quad M_{qn} l_n \hat{q}_n] \quad (6)$$

$$M_{\omega r_q} = -M_{r_q \omega} \quad (7)$$

$$M_\omega = \begin{bmatrix} M_{c11} l_1^2 & -M_{c12} l_1 l_2 \hat{q}_1 \hat{q}_2 & \ldots & -M_{c1n} l_1 l_n \hat{q}_1 \hat{q}_n \\ -M_{c21} l_2 l_1 \hat{q}_2 \hat{q}_1 & M_{c22} l_2^2 & \ldots & -M_{c2n} l_2 l_n \hat{q}_2 \hat{q}_n \\ \vdots & \vdots & \ddots & \vdots \\ -M_{cn1} l_n l_1 \hat{q}_n \hat{q}_1 & -M_{cn2} l_n l_2 \hat{q}_n \hat{q}_2 & \ldots & M_{cnn} l_n^2 \end{bmatrix}$$

(8)

In addition to, Coriolis matrix and control input are represented as

$$C = [C_{r_q} \quad C_{\omega_1} \quad \ldots \quad C_{\omega_n}]^T \quad (9)$$

$$T = [-fRe_3 + F_{dis} \quad 0]^T \quad (10)$$

Here sub-matrices of the Coriolis matrix are

$$C_{r_q} = \sum_{i=1}^{n} M_{qi}(-l_i \| \omega_i \|^2 q_i) - M_T g e_3 \quad (11)$$

$$C_{\omega_i} = -\sum_{\substack{j=1 \\ i \neq j}}^{n} M_{cij} l_i l_j \hat{q}_i^2 \| \omega_j \|^2 q_j - M_{qi} g l_i \hat{q}_i^2 e_3 \quad (12)$$

## III. CONTROL DESIGN

To facilitate the transport of the load to its intended destination, we design a well-structured hierarchical control system. The initial segment of this system is the attitude control, functioning as an internal control loop, which ensures the quadrotor's stability based on the desired angles. Subsequently, the second part governs the translational motion of the quadrotor. This section is responsible for generating thrust force and determining the desired angles. Lastly, the navigation and obstacle avoidance control loop guarantees that the system keep to the specified path towards the desired position by moving among obstacles.

*A. Navigation and obstacle avoidance control*

To carry the load, the navigation control creates an artificial point which guarantees to do two tasks. First, path planning is done through avoiding colliding with dynamic and static obstacles during the transportation of load in an environment with a lot of obstacles. Second, it picks the shortest path by tuning the forces. In this research, our objective is to determine the optimal path for transporting the load to its desired location. This is achieved by employing an enhanced AFP (Artificial Potential Field) approach in conjunction with Particle Swarm Optimization (PSO) algorithms. In essence, PSO is used to minimize a cost function, which in turn optimizes the repulsive attraction and damping forces associated with the improved AFP.

In the following, it is explained how the navigation and obstacle avoidance control of the system is done through using artificial potential field (APF).

*1) Improved Artificial Potential Field*

The Improved Artificial Potential Field (IAPF), originally introduced in [19], operates by applying attractive forces from the target point and repulsive forces from nearby obstacles. The attractive potential field is mathematically defined as follows:

$$U_{att} = \frac{1}{2} k_t \rho^2(\boldsymbol{r_l}, \boldsymbol{r_t}) \tag{13}$$

where $k_t = diag([k_{xt}, k_{yt}, k_{zt}]) \in \mathbb{R}^{3 \times 3}$ represents the attractive gain and $\rho^2(\boldsymbol{r_l}, \boldsymbol{r_t}) = (r_{xl} - r_{xt})^2 + (r_{yl} - r_{yt})^2 + (r_{zl} - r_{zt})^2$ represents the distance between the target and load position, and $\boldsymbol{r_l} = \boldsymbol{r_q} - nL\boldsymbol{e_3}$ represents the difference between the quadrotor's position and the cable length. The gradient of the potential field translates into the gravitational force

$$F_{att}(\boldsymbol{r_l}) = -\nabla U_{att}(\boldsymbol{r_l}) = -\boldsymbol{k_t}(\boldsymbol{r_l} - \boldsymbol{r_t}). \tag{14}$$

The repulsive potential field is expressed as:

$$U_{rep}(\boldsymbol{r_l}) = \begin{cases} \frac{1}{2} k_m (\frac{1}{\rho(r_l, r_{obs})} - \frac{1}{\rho_0})^2 (\boldsymbol{r_l} - \boldsymbol{r_t})^n & \text{if } \rho(\boldsymbol{r_l}, \boldsymbol{r_{obs}}) \leq \rho_0 \\ 0 & \text{if } \rho(\boldsymbol{r_l}, \boldsymbol{r_{obs}}) > \rho_0 \end{cases} \tag{15}$$

where $\boldsymbol{r_{obs}}$, $\rho_0$, and $\boldsymbol{k_m} = diag([k_{xm}, k_{ym}, k_{zm}]) \in \mathbb{R}^{3 \times 3}$ denote, respectively, the current obstacle position, the farthest distance at which an obstacle has an effect, the closest distance between the load and the obstacle, and the damping (repulsive) gain, respectively. The parameter n takes values from 0, 0.5, 1, and 2. The repulsive force is computed by taking the gradient of the repulsive potential field as:

$$F_{rep}(\boldsymbol{r_l}) = -\nabla U_{rep}(\boldsymbol{r_l}) = \begin{cases} F_{rep1}(\boldsymbol{r_l}) + F_{rep2}(\boldsymbol{r_l}) & \text{if } \rho(\boldsymbol{r_l}, \boldsymbol{r_{obs}}) \leq \rho_0 \\ 0 & \text{if } \rho(\boldsymbol{r_l}, \boldsymbol{r_{obs}}) > \rho_0 \end{cases} \tag{16}$$

where

$$F_{rep1} = -\boldsymbol{k_m} \left(\frac{1}{\rho(r_l, r_{obs})} - \frac{1}{\rho_0}\right) \frac{1}{\rho^2(r_l, r_{obs})}$$

$$(\boldsymbol{r_l} - \boldsymbol{r_t})^n \nabla \rho(r_l, r_{obs}) \tag{17}$$

$$F_{rep2} = -\frac{n}{2} \boldsymbol{k_m} \left(\frac{1}{\rho(r_l, r_{obs})} - \frac{1}{\rho_0}\right)^2 \frac{1}{\rho^2(r_l, r_{obs})}$$

$$(\boldsymbol{r_l} - \boldsymbol{r_t})^{n-1} \nabla (\boldsymbol{r_l} - \boldsymbol{r_t}) \tag{18}$$

Ultimately, the total force during navigation $F_{total} = F_{att}(\boldsymbol{r_l}) + F_{rep}(\boldsymbol{r_l})$ and the acceleration of imaginary point is

$$\ddot{\boldsymbol{r}}_p = F_{total}. \tag{19}$$

*2) Particle Swarm optimization*

Particle Swarm Optimization (PSO) stands as a highly regarded metaheuristic optimization algorithm, first introduced by James Kennedy and Russell Eberhart in 1995 [20]. It has a broad application in different areas such robotics [21], image processing [22], economics [23], and many other fields. Its primary purpose is to find the most suitable solution for optimization problems. PSO boasts several advantages, including straightforward implementation, minimal algorithm parameters, and high efficiency in global search tasks. PSO takes inspiration from the social and cooperative behaviours observed in various species when seeking resources in their environment. The algorithm considers personal experience ($pbest$), overall experience ($gbest$), and the current movement of particles to determine their next positions within the search space. Initially, a population (swarm) of size N with dimension D is represented as $X = [X_1, X_2, ..., X_N]^T$ where $X_i$ ($i = 1, 2, ..., N$) is given as $X_i = [X_{i1}, X_{i2}, ..., X_{iD}]$. Moreover, the initial velocity of the population is denoted as $V = [V_1, V_2, ..., V_N]^T$. Thus, the velocity of the ith particle is expressed as $V_i = [V_{i1}; V_{i2}; ...; V_{iD}]$.

The particle updates the position and velocity according to the following equation:

$$V_{ij}(k+1) = w(k)V_{ij}(k) + c_1(k) r_1 \left(pbest_{ij} - X_{ij}(k)\right)$$

$$+ c_2(k) r_2 \left(gbest_{ij} - X_{ij}(k)\right) \tag{20}$$

$$X(k+1) = X_{ij}(k) + V_{ij}(k). \tag{21}$$

Here, $i$ varies from 1 to $N$ and $j$ varies from 1 to $D$; $k$ denotes the iteration number, and the experiences are influenced by two factors, $c_1$ and $c_2$, while $r_1$ and $r_2$ represent random numbers generated within the range [0,1]. The current movement is scaled by an inertia weight, $w$. The implimention of the algorithm is explained by Fig. 2. However, in its standard form, constants are used for $w$ and acceleration variables ($c_1, c_2$), which may hinder convergence. It is often beneficial to have larger values for these factors during the initial convergence stages and smaller values, particularly when particles are near the optimal solution, for improved optimization results. Consequently, constructing $w$ as a decay function is essential for achieving an optimal global path during the search. In this work, three different types of PSOs are applied to investigate the convergence and minimizing the error: the classic PSO (constant coefficients), the time varying inertia weight PSO (PSO-TVIW) and the self-adaptive PSO (SAPSO). The inertia weight, $c_1$ and $c_2$ are define in these cases, respectively, by:

1. Classic PSO: $w = const., c_1 = const., c_2 = const.$
2. PSO-TVIW: $w = w_{max} - (w_{max} - w_{min}) \cos\left(\frac{\pi k}{2N}\right)$, $c_1 = const., c_2 = const.$

3. SAPSO  $w = 1 - e^{-\frac{\bar{V}(k)}{V_{max}}}$ , $c_1(k) = \alpha\left(1 - e^{-\frac{\bar{V}(k)}{V_{max}}}\right)$, $c_2(k) = \alpha - c_1(k)$,

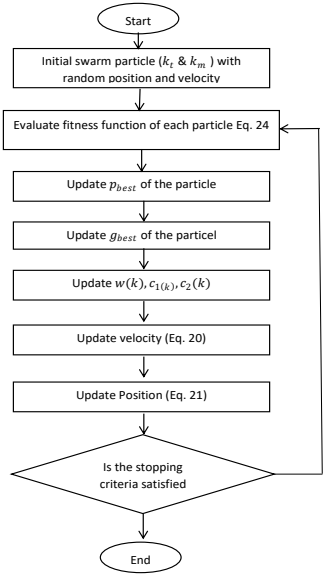

Fig 2: PSO algorithm flowchart.

where $\alpha > 0$, the $w_{max}$ and $w_{min}$ are, respectively, the upper and lower bound of the inertia weight, and k is the iteration. Additionally,

$$\bar{V}(k) = \frac{1}{ND}\sum_{i,j}^{N,D}|v_{ij}(k)| \qquad (22)$$

$$V_{max} = \frac{1}{D}\sum_{j=1}^{D}|v_{jmax}|. \qquad (23)$$

In comparison with the classic PSO, PSO-TVIW and SAPSO because of adaptive inertia weight and acceleration coefficients can improves the abilities of searching for the global optima, speed up the convergence rate and reduce the total iterations.

In the context of the PSO algorithm, the precise formulation of the fitness function plays a pivotal role in the quest for the perfect optimal solution. In this study, our objective is to identify the most optimal path from a multitude of potential paths, with the aim of minimizing the discrepancy between the desired and current positions of the load, along with reducing the time it takes to complete the task. Consequently, the cost or fitness function is defined as the integral of the product of the error's magnitude and the time duration, expressed as follows:

$$J(e,t) = \int t\|e\|dt \qquad (24)$$

where the norm of error is defined as following:

$$\|e\| = \sqrt{e_{xl}^2 + e_{yl}^2 + e_{zl}^2}. \qquad (25)$$

By finding the minimum value of the fitness function, the coefficient of repulsive and attractive forces $[k_{mx}, k_{my}, k_{mz}, k_{tx}, k_{ty}, k_{tz}]^T$ are tuned.

### B. Position Control

The design of quadrotor position control is aimed at regulating the quadrotor's position based on a reference point integrated into the navigation loop. Given the existence of external disturbances such as wind disturbances and unmodeled dynamics, the sliding mode approach, which is used in such chaotic and unknown situations [5] and [16], is employed for position control. The general dynamic position of the quadrotor is formulated as following:

$$\begin{cases} \dot{r}_q = V \\ \ddot{r}_q = M^{-1}[U + \Delta - F_p] \end{cases} \qquad (26)$$

which here $U = [u_x, u_y, u_z] \in \mathbb{R}^3$ $\Delta = [\delta_x, \delta_y, \delta_z] \in \mathbb{R}^3$, and $F_p = [F_{px}, F_{py}, F_{pz}] \in \mathbb{R}^3$ represent control inputs, the disturbances, and load effects, respectively. The state errors of the quadrotor are defined by as

$$e = r_p - r_q \qquad (27)$$

$$e_v = \dot{r}_p - \dot{r}_q \qquad (28)$$

such that $r_p$ and $\dot{r}_p$ are respectively the position and velocity of imaginary point. Hence, the sliding surface is given by following definition

$$S = \Lambda e + e_v. \qquad (29)$$

In here $S = [s_x \ s_y \ s_z]^T \in \mathbb{R}^3$ and $\Lambda = diag(\lambda_x, \lambda_y, \lambda_z) \in \mathbb{R}^{3\times 3}$ where $\lambda_x, \lambda_y, \lambda_z > 0$. By differentiating (29), we have

$$\dot{S} = \Lambda e_v + \dot{e}_v = 0 \qquad (30)$$

Finally, the control signal $U$ can be determined as

$$U = M\{\hat{f}_p - \Lambda e_v - f_d + \ddot{r}_p - M sgn(s)\} \qquad (31)$$

where $M = diag(\mu_x, \mu_y, \mu_z) \in \mathbb{R}^{3\times 3}$ and the parameter $f_d = [f_{dx}, f_{dy}, f_{dz}]^T \in \mathbb{R}^3$ denote the upper bound of $\Delta$ and the function $sgn(s) = [sgn(x), sgn(y), sgn(z)]^T \in \mathbb{R}^3$.

***Theorem 1***. The close-loop control system with the dynamics (26), sliding surface (29), and control input (31) is asymptotically stable, if $\mu_x > f'_{dx} + f'_{px} + \eta_x$.

***Proof***: a detailed proof was presented at [14]. ∎

### C. Attitude Control

The purpose of attitude control is to ensure the stability of the quadrotor's rotational motion. A PID (Proportional-Integral-Derivative) approach is employed to achieve stabilization of the quadrotor's angles, as outlined below:

$$U_2 = k_{p\phi}(\phi_d - \phi) - k_{d\phi}\dot{\phi} + k_{i\phi}\int(\phi_d - \phi) \qquad (44)$$

$$U_3 = k_{p\theta}(\theta_d - \theta) - k_{d\theta}\dot{\theta} + k_{i\theta}\int(\theta_d - \theta) \qquad (45)$$

$$U_4 = k_{p\psi}(\psi_d - \psi) - k_{d\psi}\dot{\psi} + k_{i\psi}\int(\psi_d - \psi). \qquad (46)$$

In these equations $\varphi_d, \theta_d$ and $\psi_d$ (that $\psi_d = 0$) represent the quadrotor desired angles respect to its body frame; $\varphi, \theta$ and $\psi$ are controlled through the proportional coefficients $k_{p\varphi}, k_{p\theta}$ and $k_{p\psi}$, respectively; the derivative coefficients for control of the angular velocities are shown by $k_{d\varphi}, k_{d\theta}$ and $k_{d\psi}; \dot{\varphi}, \dot{\theta}$ and $\dot{\psi}$ are the quadrotor velocity; finally, $k_{i\varphi}, k_{i\theta}$ and $k_{i\psi}$ are the integrator coefficients. Additionally, the desired angular velocities of the quadrotor are supposed zero.

## IV. SIMULATION RESULTS

In this section, the performance of optimal trajectory control strategy for quadrotor suspended-payload are

validated by numerical results. The properties of the dynamics model are given by

$$n = 3, m_3 = 0.25kg, m_{1,2} = 0.05kg, \quad L_i = 0.25m,$$
$$m_q = 0.775kg, J_q = diag(0.577, 0.577, 1.05) \times 10^{-2} \frac{kg}{m^2}.$$

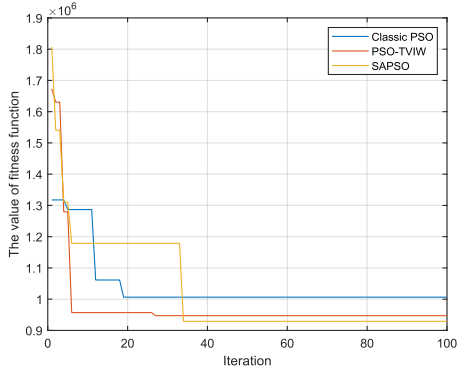

Fig 3: Fitness Function Convergence of PSO Algorithm.

The desired location and velocity of the load are $[45, 60, -10]^T$ and $[0,0,0]^T$. Additionally, control parameters are selected as

$k_p = diag(0.05, 0.05, 0.08)$, $k_d = diag(0.6, 0.6, 0.8)$, $k_i = diag(0.15, 0.15, 0.1)$, $\Lambda = diag(0.04, 0.04, 0.8)$, $M = diag(0.06, 0.06, 0.08)$.

Moreover, the coefficient of repulsive and attractive force are selected as following

$k_{xm}, k_{ym}, k_{zm} \in [0.001, 1].,\quad k_{xt} k_{yt}, k_{zt} \in [0.001, 1]$.

Also, in classic PSO $w = 0.6$ and in PSO_TVIW and SAPSO $w \in [0.5, 0.9]$, meanwhile, $c_1, c_2 = 2$. The number of particles is 50 and the number of iterations is 100. In here, the static obstacles are randomly dispersed and moving obstacles are assume to have a random velocity between [0.001, 0.5].

The SAPSO, PSO-TVIW, and classic PSO optimization algorithms generate the values for the repulsive and attractive coefficients, respectively as

$k_m = diag([0.0649, 0646, 065])$ and $k_{tx} = diag([0.0122, 0.0121, 0.0123])$, $k_m = diag([0.0656, 0.0654, 0.0653])$ and $k_t = diag([0.0146, 0.0146, 0.014])$, $k_m = diag([0.0622, 0.0625, 0.0624])$ and $k_t = diag([0.0086, 0.0085, 0.0086])$.

As it is shown in Fig. 3, the value of fitness function for classic PSO is higher than two other methods since the other PSO approaches, particularly SAPSO, has the ability of searching for the global optima.

Additionally, Figures 4 indicates the proper performance of the controller such that it finds and tracks the closest path in the environment with many number of fixed and moving obstacles. Indeed, all three methods generate the optimal paths, but the generated trajectory by the SAPSO, it moves closer to obstacle and reduce the distance to the target position(error). Although during the operation, the load has some fluctuations, but the control system compensated them as well, Fig. 4(a) and Fig. 5.

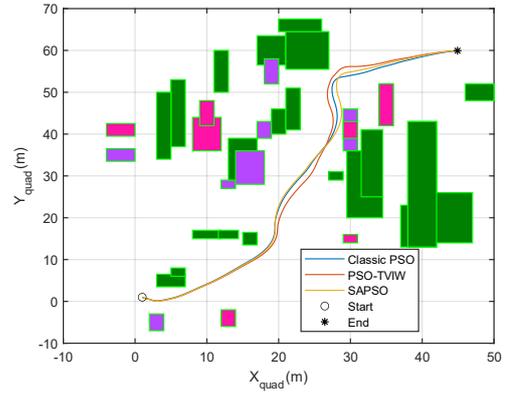

(a)

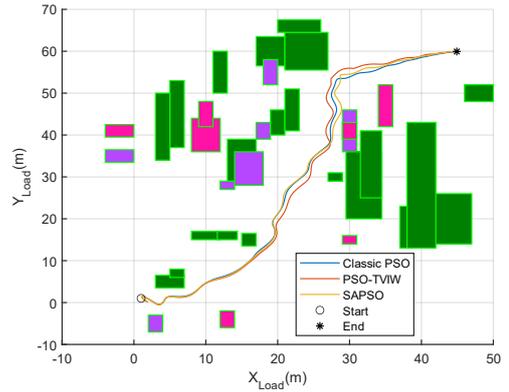

(b)

Fig 4: Operation of load transportation in x-y plan, a) Quadrotor position and b) Load postion. Great objects are fixed, pink rectangles are the initial position of moving obstacles and purple rectangles are final condition.

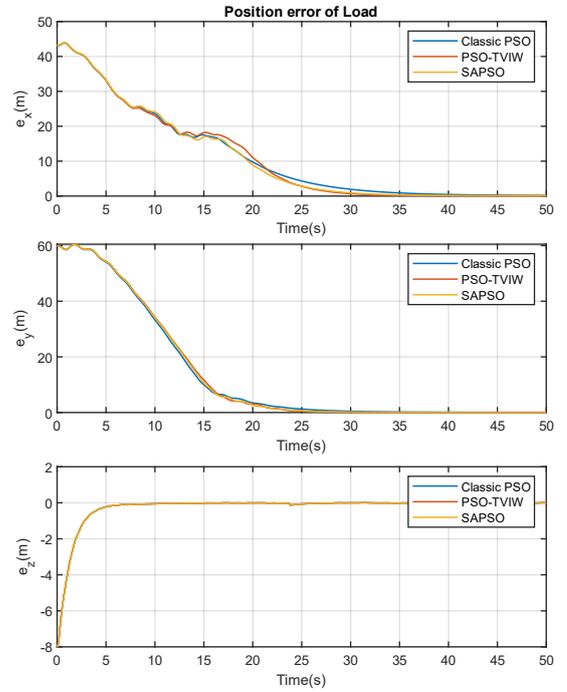

Fig 5: Error of load position in x direction, y-direction, and z direction.

In Fig. 5-6, it is indicated that since the obstacle to be consider with a huge height so in z direction system converge to desired z position faster. Generally speaking, the settling time of SAPSO and PSO-TVIW is less than classic PSO.

## V. Conclusion

This article explores the optimization of flight paths for transporting loads using drones in both busy and stable environments. The study focuses on utilizing Particle Swarm Optimization (PSO) for this purpose. A hierarchical control system was developed for a quadrotor equipped with a cable-suspended payload, and its motion equations were derived using the Euler-Lagrange approach. To navigate the drone safely through obstacles, a combination of an enhanced artificial potential field and the PSO algorithm was employed. An abstract reference point, acting as a guiding leader, was utilized to guide the system towards its target while avoiding collisions with static and moving obstacles. The gravitation and repulsion forces' coefficients were adjusted using various PSO methods to achieve the optimal trajectory and minimize the time required for transport. This reference point was crucial for precise position control of the quadrotor, employing a sliding mode strategy. Ultimately, the numerical results were presented to validate the effectiveness of the system in successfully transporting the payload.

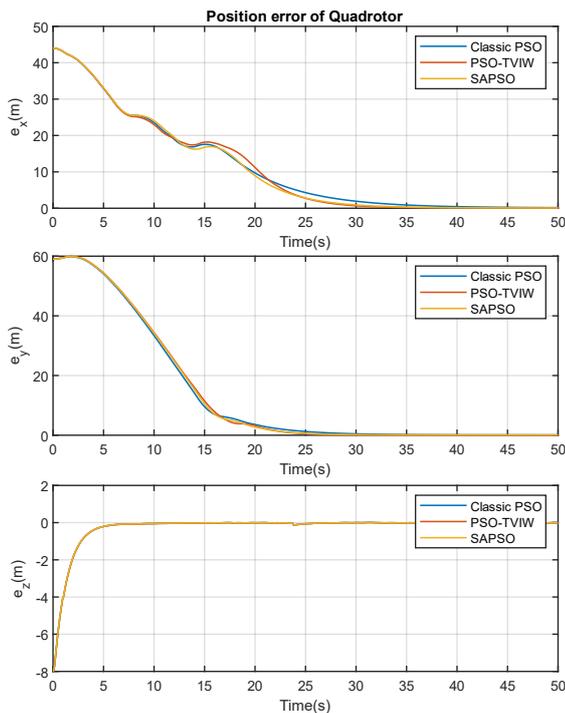

Fig 6: Error of quadrotor position in x direction, y-direction, and z direction.

## References

[1] K. Tom, and S. Jaroslaw, Quadrotor UAV control for transportation of cable suspended payload, ser. Lecture Notes in Statistics. Journal of KONES, vol. 26, No, 2, pp. 77–84, 2019.

[2] S. Sadr, and S. A. Moosavian, "Dynamics modeling and control of a quadrotor with swing load," Journal of Robotics, 2014.

[3] B. Lee, S. Hong, D. Yoo, and H. Lee, "Design of a neural network controller for a slung-load system lifted by 1 quad-rotor," Journal of Automation and Control Engineering, vol. 3, No. 1, pp. 9-14, 2015.

[4] P. J. Cruz, and R. Fierro, "Cable-suspended load lifting by a quadrotor UAV: hybrid model, trajectory generation, and control," Autonomous Robots, Vol. 41, No. 8, 2017, pp. 1629—1643.

[5] Yaghooti, B., Safavigerdini, K., Hajiloo, R. and Salarieh, H., 2022. Stabilizing unstable periodic orbit of unknown fractional-order systems via adaptive delayed feedback control. arXiv preprint arXiv:2208.06783.

[6] P. Kotaru, G. Wu and K. Sreenath, "Dynamics and control of a quadrotor with a payload suspended through an elastic cable," 2017 American Control Conference (ACC), Seattle, WA, USA, 2017, pp. 3906-3913, doi: 10.23919/ACC.2017.7963553.

[7] M. Gue, Y. Su and D. Gu, "Mixed H2/H∞ tracking control with constraints for single quadcopter carrying a cable-suspended Payload", IFAC-PapersOnline, vol. 50, no. 1, pp. 4869-4874, 2017.

[8] X. Liang, Y. Fang, N. Sun and H. Lin, "Dynamics analysis and time optimal motion planning for unmanned quadrotor transportation systems," Mechatronics, vol. 50, pp. 16-29, 2018.

[9] I. Palunko, P. Cruz and R. Fierro, "Agile load transportation: safe and efficient load manipulation with aerial robots," IEEE Robotics Automation Magazine, vol. 19, pp. 69-79, 2012.

[10] J. G. Romeroa and H. Rodríguez-Cortés, "Asymptotic stability for a transformed nonlinear UAV model with a suspended load via energy shaping," European Journal of Control, vol. 52, pp. 87-96, 2020.

[11] D. Hashemi and H. Heidari, "Trajectory planning of quadrotor UAV with maximum payload and minimum oscillation of suspended load using optimal control." J. Intell. Robot. Syst., 2020.

[12] A. A. R. Lori, M. Danesh, P. Amiri, S. Y. Ashkoofaraz, and M. A. Azargoon, "Transportation of an Unknown Cable- Suspended Payload by a Quadrotor in Windy Environment under Aerodynamics Effects," 2021 7th International Conference on Control, Instrumentation and Automation (ICCIA), IEEE, 2021, pp. 1–6.

[13] Hoorfar H, Kosarirad H, Taheri N, Fathi F, Bagheri A. Concealing Robots in Environments: Enhancing Navigation and Privacy through Stealth Integration. EAI Endorsed Transactions on AI and Robotics. 2023.

[14] Hoorfar, H., Fathi, F., Moshtaghi Largani, S., Bagheri, A. Securing Pathways with Orthogonal Robots. The 21st International Conference on Scientific Computing, Las Vegas, USA, 2023.

[15] M. Kargar, C. Zhang and X. Song, "Integrated Optimization of Power Management and Vehicle Motion Control for Autonomous Hybrid Electric Vehicles," in IEEE Transactions on Vehicular Technology, vol. 72, no. 9, pp. 11147-11155, Sept. 2023, doi: 10.1109/TVT.2023.3270127.

[16] S. P. H. Boroujeni,, & A. Razi,. "IC-GAN: An Improved Conditional Generative Adversarial Network for RGB-to-IR image translation with applications to forest fire monitoring". Expert Systems with Applications (2023),121962.

[17] C. Y. Son, H. Seo, D. Jang and H. J. Kim, "Real-Time Optimal Trajectory Generation and Control of a Multi-Rotor With a Suspended Load for Obstacle Avoidance," in IEEE Robotics and Automation Letters, vol. 5, no. 2, pp. 1915-1922, 2020.

[18] S. Y. Ashkoofaraz and A. A. R. Lori, "Aerial Load Transportation With Obstacle Avoidance in Observed Environment," 2022 10th RSI International Conference on Robotics and Mechatronics (ICRoM), Tehran, Iran, 2022, pp. 248-253.

[19] J. Sun, J. Tang and S. Lao, "Collision Avoidance for Cooperative UAVs With Optimized Artificial Potential Field Algorithm," in IEEE Access, vol. 5, pp. 18382-18390, 2017.

[20] R. Eberhart, J. Kennedy, ''A new optimizer using particle swarm theory, in: In: MHS'95. Proceedings of the Sixth International Symposium on Micro Machine and Human Science, 1995, pp. 39–43.

[21] S. Mohamad Ali Tousi, A. Mostafanasab and M. Teshnehlab, "Design of Self Tuning PID Controller Based on Competitional PSO," 2020 4th Conference on Swarm Intelligence and Evolutionary Computation (CSIEC), Mashhad, Iran, 2020, pp. 022-026.

[22] N. Mehrabi and S. P. H. Boroujeni, "Age Estimation Based on Facial Images Using Hybrid Features and Particle Swarm Optimization," 2021 11th International Conference on Computer Engineering and Knowledge (ICCKE), Mashhazd, Iran, 2021, pp. 412-418.

[23] M.A. Abido, Environmental/economic power dispatch using multiobjective evolutionary algorithms, IEEE Trans. Power Syst. 18 (4) (Nov. 2003) 1529–1537,